\documentclass[
]{ceurart}

\usepackage{listings}
\begin{document}

\copyrightyear{2026}
\copyrightclause{Copyright for this paper by its authors.
  Use permitted under Creative Commons License Attribution 4.0
  International (CC BY 4.0).}

\conference{SEPLN 2026: 42\textsuperscript{nd} International Conference of the Spanish Society for Natural Language Processing, León, Spain, 22-25 September 2026.}

\title{CRITICS - Critical Science Without Borders: Language Models to
Promote Critical Thinking in Science Education}


\author[1]{Rodrigo Agerri}[%
email=rodrigo.agerri@ehu.eus,
]
\cormark[1] 

\author[1]{Itziar Aldabe}[%
email=itziar.aldabe@ehu.eus,
]

\author[2]{Elena Cabrio}[%
email=elena.cabrio@univ-cotedazur.fr,
]

\author[3]{Mark Cieliebak}[%
email=mark.cieliebak@zhaw.ch,
]

\author[3]{Jan Deriu}[%
email=janmilan.deriu@zhaw.ch,
]

\author[1]{Mariana Flores}[%
email=mariana.flores@ehu.eus,
]

\author[4]{Jurgita Kapočiūtė-Dzikienė}[%
email=jurgita.kapociute-dzikiene@vdu.lt,
]

\author[4]{Dovilė Kuizinienė}[%
email=dovile.kuiziniene@vdu.lt,
]

\author[5]{Arantza Rico}[%
email=arantza.rico@ehu.eus,
]

\author[5]{Aritz Ruiz-Gonz\'alez}[%
email=aritz.ruiz@ehu.eus,
]

\author[1]{Aitor Soroa}[%
email=a.soroa@ehu.eus,
]

\author[4]{Mantas Vaškevičius}[%
email=mantas.vaskevicius@vdu.lt,
]

\author[2]{Serena Villata}[%
email=serena.villata@univ-cotedazur.fr,
]

\address[1]{HiTZ Center - Ixa, University of the Basque Country EHU}

\address[2]{Université Côte d’Azur, Inria, CNRS, I3S, France}

\address[3]{Centre for Artificial Intelligence, ZHAW School of Engineering}

\address[4]{Faculty of Informatics, Vytautas Magnus University, Lithuania}

\address[5]{Department of Mathematics, Experimental and Social Sciences Education, University of the Basque Country EHU}

\cortext[1]{Corresponding author.}

\begin{abstract}
  The CRITICS project addresses science accessibility and literacy through the convergence of advanced Machine Translation (MT) based on Large Language Models (LLMs) and educational technology. By leveraging MT systems specifically optimized for scientific content, educational institutions can provide accurate, culturally relevant translations of scientific materials in students' native languages, ensuring that complex scientific concepts are comprehensible while maintaining technical accuracy. 
Building on these translations, the project explores the design and evaluation of innovative science teaching–learning proposals grounded in curriculum-aligned teaching–learning. Thus, CRITICS will investigate key components of scientific argumentation and critical thinking practices together with textual feedback aligned with learning objectives and assessment criteria inspired by competence-based evaluation frameworks. CRITICS aims to break down language barriers in accessing cutting-edge research and educational materials currently only available in high-resourced languages, thereby facilitating the democratization of scientific knowledge while fostering critical thinking in science education.
\end{abstract}


\maketitle

\section{Introduction}

CRITICS\footnote{\url{https://www.hitz.eus/critics}} is a three-year CHIST-ERA IV Cofund 2025 project funded within the topic ``Science in your own language''.\footnote{https://www.chistera.eu/projects-call-2025} This call addresses the translation of scientific knowledge to bridge linguistic and cultural gaps for those who must disseminate and access scientific knowledge beyond their linguistic scope.

The project is coordinated by the HiTZ Center from the University of the Basque Country EHU, funded by MICIU/AEI /10.13039/501100011033 and by the European Union (PCI2025-167239-2). In addition to HiTZ, the EHU team includes researchers from the Department of Mathematics, Experimental and Social Sciences Education. The consortium is also formed by the CNRS/Université Côte d’Azur in France (grant no ANR-25-CHR4-0002-02), Vytautas Magnus University (VMU) (funded by the Research Council of Lithuania, agreement No. S-CHIST-ERA-26-1), and the ZHAW Center for Artificial Intelligence in Switzerland (grant no. 20CH-1\_238349).

Scientific literacy and the development of critical thinking are essential components for a democratic society. However, access to scientific knowledge continues to be conditioned by language, since a large proportion of research articles and advanced educational resources are published almost exclusively in English or other high-resource languages such as Spanish. This situation creates linguistic inequality that limits learning opportunities for both students and teachers of science, particularly in low-resource languages such as Basque, and hinders students’ ability to learn how to construct evidence-based arguments in their own language. The CRITICS project addresses these challenges through the convergence of advanced MT systems based on language models (LLMs) and educational technologies.

The accessibility to scientific content in our own languages through advanced MT naturally connects to the automated generation of science education materials, where LLMs can be applied to synthesize and adapt complex scientific concepts into level-appropriate and pedagogically sound resources. Thus, CRITICS will develop and adapt LLMs to facilitate the generation of customized science education materials \cite{meheut2004teaching}.

CRITICS will mostly follow recent science education research focused on Design-Based Research \cite{ruiz2025learning} and consider education materials to specify the \emph{driving problem/questions}, the learning objectives focused on competency acquisition, the scientific practices including scientific argumentation and critical thinking, and the activities to be made by the science students. The availability of machine-translated scientific knowledge will be crucial to investigating and developing LLMs for the automatic generation of appropriate education materials in the students’ native languages that relate to local students' experiences \cite{ruiz2025learning}.

\begin{figure*}
  \centering
  \includegraphics[scale=0.45]{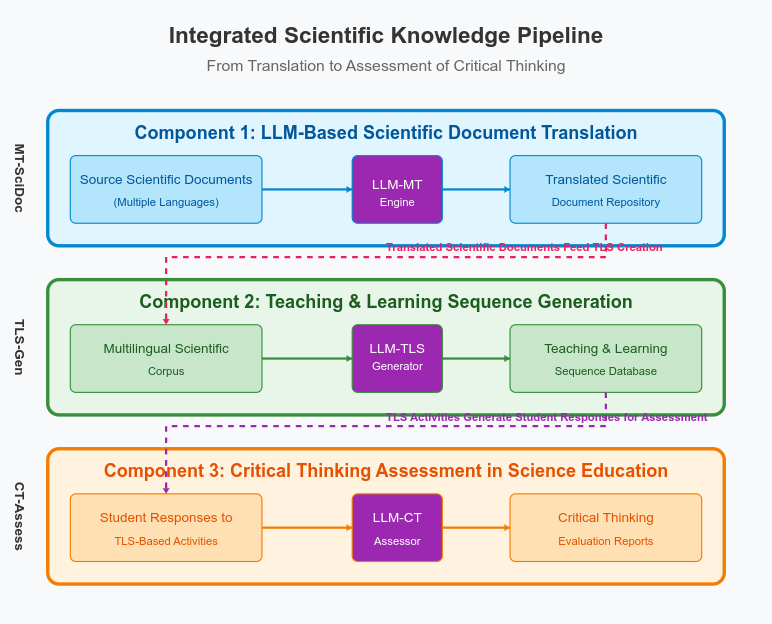}
  \caption{Three main components in the CRITICS approach.}\label{fig:pipeline}
\end{figure*}

\noindent \textbf{Overall Approach:} Figure \ref{fig:pipeline} illustrates the integrated approach provided by CRITICS, which directly addresses the core objectives of the call by creating a comprehensive ecosystem that not only machine translates scientific content but transforms it into educationally meaningful resources with built-in assessment capabilities. CRITICS will develop a three-component system as in Figure \ref{fig:pipeline} for cross-lingual scientific knowledge transfer. The \textbf{first component is an LLM-based scientific document translation system} that surpasses conventional MT approaches, which often struggle with domain-specific terminology and notation \cite{kleidermacher2026science,kim2024efficient}, using adapters fine-tuned on specialized scientific corpora to preserve structural elements and field-specific terminology. The \textbf{second component, a Teaching and Learning Sequence (TLS) generator}, will investigate how to transform translated scientific literature into structured educational resources by analyzing the corpus to identify key concepts and pedagogical approaches, extracting argumentation graphs to generate TLSs. The \textbf{third component, an LLM-based automatic assessor}, will focus on how to evaluate student responses to TLS-derived activities by analyzing evidence-based argumentation and applying automatically generated scientific critical questions, with Argument Mining capabilities to detect misconceptions and fallacies. The CRITICS LLMs adapted to the tasks will be placed in a continuous improvement loop where assessments feed back into translation quality, teaching resource development, and assessment criteria refinement.

\section{Related Work and Novelty}\label{sec:related-work}

CRITICS will comply with TRL 5 – technology validated in relevant environments (industrially relevant environment in the case of key enabling technologies),\footnote{https://horizoneuropencpportal.eu/sites/default/files/2022-12/trl-assessment-tool-guide-final.pdf} where the qualitative evaluation will be done involving domain-experts (science education teachers and researchers on the two main disciplines targeted in the project, natural science and AI) in both the supervision and the actual evaluation.

\subsection{LLM-based MT of Scientific Documents}

LLMs have substantially advanced the field of MT, particularly for high-resource languages and general-purpose domains. Proprietary models deliver strong fluency and document-level coherence, benefiting from long-context modeling and multilingual pretraining, though notable gaps remain in low-resource scenarios \cite{wang2023document,zhu2024multilingual}. Prompt-based strategies (especially zero- and few-shot prompting) have emerged as effective methods for improving LLM translation performance \cite{zhang2023prompting}. Terminology-aware prompting and term injection have received growing attention as key techniques for addressing a persistent challenge in domain-specific MT: terminological inconsistency. This is especially critical in scholarly translation, where inaccurate terminology can distort meaning or undermine academic precision \cite{kleidermacher2026science}. Recent work highlights diverse strategies for integrating terminology into LLM-based MT \cite{sabo2024boosting,lu2024chain,kim2024efficient}. Despite these advances, less-resourced languages have received limited attention in LLM-based MT research. Recent work \cite{kapovciute2025localizing} demonstrates promising results for English–Lithuanian translation using open-weight LLMs. However, these findings are based on benchmark data rather than real scholarly texts, thus highlighting the need for more work in domain-specific, low-resource scenarios.\\

\noindent \textbf{Progress beyond the state-of-the-art:} This project builds on recent advances in LLM-based MT but goes further by addressing the core challenges of scholarly translation (particularly for low-resource languages such as Lithuanian and Basque, with French included for comparative purposes). Our focus extends beyond fluency and general translation quality to include terminological precision, structural accuracy, and document-level coherence – i.e., key requirements in scholarly texts.
We will benchmark a range of available LLMs, prioritizing open-weight models with good performance on the target languages such as Gemma \cite{team2024gemma}, Llama \cite{grattafiori2024llama} or language-specific models such as Latxa \cite{etxaniz2024latxa}, to identify the most effective options for our target language pairs: English–Lithuanian, English–Basque, English–French and English-German. A central focus of the project is document-level translation, where maintaining consistency of terminology and coherence across sections is essential. To support this, we will develop domain-specific terminology glossaries. These glossaries will be integrated into the translation workflow through prompt engineering or as part of the fine-tuning datasets. Beyond linguistic accuracy, we will also develop tools to preserve the layout and structure of scholarly documents, ensuring that translations retain citations, equations, code, and section formatting.
\subsection{Argument(ation) Mining (AM)} 

Existing research on mining, assessing, reasoning over, and generating arguments \cite{stede2019argumentation,lawrence2019argument} involves the automatic extraction of arguments and their relationships from text. Starting from the identification of claims, premises, and evidence in a piece of text, as well as the relationships between them, such as support, attack, or rebuttal, computational argumentation methods can be used to analyze the strengths and weaknesses of arguments, their quality, and to identify common patterns and themes. Usually, supervised learning methods are used to address AM tasks, leading to the need to define beforehand annotated datasets for the specific tasks and application scenarios, even if recently some empirical approaches have been proposed to synthesize and generate arguments. Among the recent challenges of the AM field (Dagstuhl Seminar 22432 “Towards a Unified Model of Scholarly Argumentation”), there is the task of detecting manipulation in scientific publications through the identification of duplicate arguments and the computation of argument similarity \cite{glockner2024missci}, which strictly relates to the fallacy detection task. Moreover, a few works have focused on highlighting distortions in scientific communication \cite{augenstein2021determining}, as well as in detecting quantified information mismatches between reported and actual scientific findings \cite{wright2022modeling}.\\

\noindent \textbf{Progress beyond the state-of-the-art:} While scientific texts aim to present verifiable evidence for a series of stated claims, the interpretation of such objective sources of evidence is often ambiguous and subjective. In this project, we will first enhance and tailor Argument mining approaches to address domain-specific features of scientific argumentation in science education texts. Then, we will explore the notion of similarity between natural language arguments, which is currently unexplored, but it would be crucial to check the validity and coherence of argumentative scientific content across languages, both for human-generated and machine-translated texts. The result of this task is the automatic extraction of the full argument graph, on which further inferences can be applied. Such argumentation graphs will also be exploited for critical questions generation, to evaluate critical thinking in scientific contexts across languages. Applications of argument mining in scientific discourse have been so far limited, in particular, because current methods are insensitive to the factual content of scientific arguments, an issue that will be carefully addressed in the context of CRITICS, following our work in the medical education domain \cite{molinet2024explanatory}.

\subsection{Critical Questions Generation and Automatic Assessment} 

Critical questions are vital for competency-based science assessments because they address deeper understanding beyond fact memorization to apply scientific knowledge and skills in practical contexts, which is the central aim of competency-based assessment. Unlike simple recall questions, critical questions help to evaluate essential higher-order thinking skills, measure students' ability to transfer knowledge to new topic-specific contexts and provide diagnostic insights into reasoning patterns and misconceptions \cite{lemons2013questions,nussbaum2021critical}. There have been approaches to automated question and feedback generation \cite{gao2024automatic,fu2024qgeval}, but automatic Critical Questions Generation is a largely unexplored task \cite{calvo-figueras-agerri-2025-benchmarking}, although some works have argued about their utility in identifying fallacious reasoning, and also for assessing the quality of argumentative essays \cite{song2014applying}.\\

\noindent \textbf{Progress beyond the state-of-the-art:} Text-based automated assessment systems have been developed to automatically grade, classify or to provide feedback or guidance \cite{gao2024automatic}. However, CRITICS will provide for the first time textual feedback for students by automatically generating Critical Questions, argument feedback and grades based on the TLSs developed for  topic specific competency-based assessments in natural sciences and AI. More specifically, the generation of Critical Questions will be aligned with specific learning goals and assessment criteria specified in the TLSs, ensuring that the feedback is relevant and focused on the intended learning outcomes. Note that the generation will be multilingual, namely, in Basque, French, German and Lithuanian. These questions often apply models such as the Bloom’s taxonomy, CER (Claim, Evidence, Reasoning) or CQMAA (Critical Questions Model of Argument Assessment) to structure students' responses and evaluations \cite{nussbaum2020using,nussbaum2021critical}.

\subsection{Evaluation and Benchmarking}  

MT evaluation typically relies on automated metrics like BLEU, ChrF++, TER, or COMET. Though WMT includes some biomedical translation resources, comprehensive scientific text datasets remain notably absent \cite{kleidermacher2026science}. However, automatic evaluation of text generation tasks remains an open challenge. In CRITICS, these tasks will be focused (apart from MT, covered above) in the evaluation of Critical Questions and TLS, and Automatic Feedback Generation. Although the task of automatic Critical Questions Generation is rather new \cite{calvo-figueras-agerri-2025-benchmarking}, automatic metrics for question generation in general have been investigated. Novel approaches like QGEval offer multi-dimensional assessment \cite{fu2024qgeval}. Still, there is a strong research interest in devising more comprehensive LLM-based evaluation methods \cite{gu2024survey,li2025generation}, also for question generation in particular \cite{fu2024qgeval}. Despite these new automatic methods, human evaluation is still considered to be essential. Thus, expert assessments would focus on grammatical correctness, relevance, semantic soundness, quality of the evidence-based argumentation, and (in some cases) alignment with cognitive frameworks like Bloom's Taxonomy \cite{lawrence2019argument}. 

The consortium has ample research experience regarding evaluation and benchmarking relevant to the CRITICS project. EHU has developed a state-of-the-art open-source LLM for Basque, Latxa \cite{etxaniz2024latxa}. EHU, in collaboration with CNRS, has released the first multilingual LLM for the medical domain, evaluating it on medical and scientific benchmarks. EHU has collected the largest corpora available for Basque, which contains circa 1.4 billion tokens \cite{etxaniz2024latxa}, and has experience in the development of new benchmarks for multilingual evaluation of LLMs \cite{alonso2024medexpqa,etxaniz2024latxa,garcia2024medmt5}. Regarding Argument Mining, CNRS and EHU have released the first multilingual dataset for the medical domain annotated with argument structures \cite{sviridova2024casimedicos} and the first publicly available dataset of Critical Questions \cite{calvo-figueras-agerri-2025-benchmarking} linked to Argument Schemes. EHU has already published various results with similarity and LLM-based evaluation methods in pairwise ranking- and reference-based settings \cite{calvo-figueras-agerri-2025-benchmarking,yeginbergen2025dynamic,zubiaga2024llm}.
ZHAW has developed theoretical foundations for the applications of automated evaluation for generative models (with a focus on Machine Translation) \cite{von2024favi,deriu2023correction}. This provides a strong foundation for correctly applying automated evaluation and calibrating the metrics to avoid biased evaluation results \cite{von2024improving}. They also published work on the robustness and reliability of trained metrics \cite{deriu2023correction}.\\

\noindent \textbf{Progress beyond the state-of-the-art:} The use of LLM-based evaluation for text generation tasks related to MT and Critical Questions and Feedback generation in the scientific domain remains an open research problem. CRITICS will provide new LLM-based evaluation methods specifically tailored to the relevant features of translating MT of scientific documents and to the science teaching-related criteria of the automatically generated Critical Questions and Assessment/Feedback.

\section{Methodology and Work Plan}\label{sec:work-plan}

Although CRITICS’ vision applies across disciplines, the project focuses on two areas: (i) natural sciences (biology, chemistry, and physics) and (ii) Artificial Intelligence. The former is key to competency-based assessments, while the latter poses specific translation challenges due to the continuous introduction of new terminology that may lack established equivalents in less-resourced languages \cite{kleidermacher2026science,zhang2024sciinstruct}. CRITICS will target a diverse spectrum of languages, namely, Basque (agglutinative language isolate), Lithuanian (East Baltic, inflectional), German (West Germanic, inflectional), and French (Romance, synthetic-fusional). Every objective will be addressed in a multilingual setting, namely, including the four target languages of the project.

\begin{enumerate}
\item \textbf{Objective 1.} LLM-based Machine Translation of Scientific Documents: Focused on developing and adapting open-weight LLMs for high-quality, document-level MT of scientific texts, particularly in low- and medium-resource language pairs (e.g., English–Lithuanian/Basque/French/German).
\item \textbf{Objective 2.} Argumentation serves as a fundamental mechanism in scientific discourse, facilitating the process of reaching conclusions and facilitating science literacy and critical thinking. This objective will focus on training LLMs to recognize evidence-based argumentation and identify fallacies and scientific misconceptions.
\item \textbf{Objective 3.} Automatic Assessment and Critical Thinking: Rather than merely identifying gaps in the scientific discourse, automatic assessment will also involve the generation of Critical Questions and Detailed Feedback in competency-based assessment settings (PISA style - Programme for International Student Assessment) where cross-linguistic comparability is essential.
\item \textbf{Objective 4.} Evaluation: qualitatively evaluate the generation of critical questions and automatic assessment feedback using LLMs-as-a-Judge paradigm in a way that can be compared with human-generated judgments \cite{calvo-figueras-agerri-2025-benchmarking}.
\end{enumerate}

\subsection{Work Plan}

The project work plan consists of three Development-Evaluation cycles that will be replicated for each of the technical work packages (WP2-WP5). The overall structure of the work plan structured in work packages is as follows. In a first step, in WP2 the consortium will define the final requirements for the use cases (Parallel Translator of Scientific Documents and Teacher Evaluator) of the project. Also in WP2 the data collection required to address the objectives and research questions related to each use case will be performed. Regarding the MT use case, most of the relevant data sources for natural sciences (biology, chemistry, and physics) and Artificial Intelligence are already identified following open science criteria.

\noindent \textbf{MT of scientific data.} The choice of English data sources aims to ensure a massive amount of scientific content to develop the LLM-based MT of scientific articles in WP3 and WP5 while being able to distribute the pre-processed corpus and the scientific outcomes. Data sources will include Biorxiv\footnote{https://api.biorxiv.org/} (biology and biochemistry), PubMed Central\footnote{https://pmc.ncbi.nlm.nih.gov/} (biomedical and life sciences, some data also available in French), chemrxiv,\footnote{https://chemrxiv.org/} arxiv for Physics and Artificial Intelligence,\footnote{https://arxiv.org/} ACL Anthology Corpus (AI and NLP),\footnote{https://github.com/shauryr/ACL-anthology-corpus} and other sources relevant for the scientific disciplines targeted in CRITICS, for example, evidence-based scientific educational resources such as Scientix,\footnote{https://www.scientix.eu/} STEM learning centre,\footnote{https://www.stem.org.uk/} e-bug.\footnote{https://www.e-bug.eu/} Note that pre-print papers usually include information about their peer-reviewed version, which will prioritize to automatically translate into the targeted languages. Furthermore, most of these resources provide API access to their content in some structured format to avoid any OCR of PDF articles. A final selection and analysis of the most appropriate data sources will be done in WP2.

\noindent \textbf{Competency-based Assessments data.} We will also compile and annotate (WP2) $~$500 competency-based assessments in higher education (questionnaires with student responses and corresponding teacher’s assessments) are available in-house at UPV/EHU and will be the basis for developing and evaluating the Teacher Evaluator tool and research in WP4. This data will be fully anonymized and released following privacy and open science policies.
Once both types of data are compiled and pre-processed, WP2 will annotate argumentation graphs, critical questions, and feedback/assessment for benchmarking in WP5. Using the data for evaluation and development, WP3 will primarily focus on the advancement of SOTA by training and experimenting with LLMs for MT of scientific data. In WP4 novel research will address the quality assessment of argumentation in scientific and educational texts. WP5 will investigate new techniques to evaluate challenging text generation tasks such as MT of scientific documents, Critical Questions and Automatic Feedback Generation.  Finally, WP6 will apply every result obtained in WP2, WP3 and WP4 to develop the two tools and prototypes for a Parallel Translator of Scientific Documents and a Teacher Evaluator.

\begin{enumerate}
    \item \textbf{WP2. Data Collection:} In this WP we will address the data collection, processing, and manual annotation and evaluation for benchmarking of the various modules in WP5. The correct description of the data guarantees its traceability, contributing to the explainability and transparency of the models. We also plan to obtain new high-quality instruction and alignment datasets derived from already instructed models \cite{zhang2024sciinstruct} using the magpie approach \cite{xu2024magpie} to reduce the need for expensive human-annotated instruction data.
    \item \textbf{WP3. LLM-based MT:} Adapt and optimize MT techniques for scholarly documents, with a focus on low-resource language pairs (English-Lithuanian, English-Basque) and medium-resource pairs (English-French, English-German). Improve translation quality using open-weight LLMs \cite{bai2023qwen,team2024gemma,grattafiori2024llama}, domain-specific data \cite{zhang2024sciinstruct}, and strategies like prompt engineering, terminology integration \cite{kim2024efficient}, and lightweight fine-tuning.
    \item \textbf{WP4. Argumentation:}. Tailor Argument mining approaches to address domain-specific features of scientific argumentation in multilingual educational texts. Assess the coherence of argumentative scientific content across languages. Critical questions generation \cite{calvo-figueras-agerri-2025-benchmarking}, to enhance and evaluate students’ ability to analyze complex texts, evaluate claims and evidence, and construct well-reasoned arguments.
    \item \textbf{WP5. Evaluation and Benchmarking:} First, this WP evaluates the translated scientific texts. We investigate two approaches. First, the human evaluation, and second automated evaluation procedures. For human evaluation, the main challenge is reproducibility, as human ratings suffer from high variance; thus, we will develop protocols and guidelines to reduce this issue. The main challenges for automated evaluation are that they disagree with human feedback and tend to prefer certain systems disproportionately. Thus, the second focus of this WP is developing and calibrating automated metrics to evaluate MT and assessment generation \cite{calvo-figueras-agerri-2025-benchmarking,yeginbergen2025dynamic,zubiaga2024llm}.
\end{enumerate}

\section{Concluding Remarks}

CRITICS aims to contribute to the democratization of scientific knowledge by integrating LLM-based MT with critical thinking in science education, overcoming language barriers for low- and medium-resource languages like Basque, Lithuanian, French, and German. Through its six work packages, the project will strive to deliver optimized MT for scientific documents, argumentation mining tailored to scientific texts, automated critical feedback generation, and robust evaluation methods, all validated via applications such as the Parallel Reader and Teacher Evaluator.

By addressing key challenges in terminology consistency, document structure preservation, and multilingual argumentation assessment, CRITICS advances beyond the current state-of-the-art in domain-specific NLP and educational tech. The project's emphasis on open-weight models, synthetic data generation, and LLM-as-a-judge metrics will investigate how to ensure reproducible, transparent results that foster scientific literacy across diverse linguistic contexts. 

Future extensions could scale to additional low-resource languages and integrate emerging multimodal LLMs for visual scientific content, amplifying the project's impact on global science education.

\begin{acknowledgments}
EHU's researchers acknowledge the following MCIN/AEI/10.13039/501100011033 projects: (i) CRITICS (PCI2025-167239-2) funded by MICIU/AEI /10.13039/501100011033 and co-funded by the European Union; (ii) DeepThought (PID2024-159202OB-C21) funded by ERDF, EU, and (iii) Xixare (PID2024-159566NB-I00). CNRS/UniCA research is supported by the French National Research Agency (grant ANR-25-CHR4-0002-02).
\end{acknowledgments}

\bibliography{sample-ceur}

\begin{thebibliography}{39}
\expandafter\ifx\csname natexlab\endcsname\relax\def\natexlab#1{#1}\fi
\providecommand{\url}[1]{\texttt{#1}}
\providecommand{\href}[2]{#2}
\providecommand{\path}[1]{#1}
\providecommand{\DOIprefix}{doi:}
\providecommand{\ArXivprefix}{arXiv:}
\providecommand{\URLprefix}{URL: }
\providecommand{\Pubmedprefix}{pmid:}
\providecommand{\doi}[1]{\href{http://dx.doi.org/#1}{\path{#1}}}
\providecommand{\Pubmed}[1]{\href{pmid:#1}{\path{#1}}}
\providecommand{\bibinfo}[2]{#2}
\ifx\xfnm\relax \def\xfnm[#1]{\unskip,\space#1}\fi
\bibitem[{M{\'e}heut and Psillos(2004)}]{meheut2004teaching}
\bibinfo{author}{M.~M{\'e}heut}, \bibinfo{author}{D.~Psillos},
\newblock \bibinfo{title}{Teaching--learning sequences: aims and tools for
  science education research},
\newblock \bibinfo{journal}{International Journal of Science Education}
  \bibinfo{volume}{26} (\bibinfo{year}{2004}) \bibinfo{pages}{515--535}.
\bibitem[{Ruiz-Gonz{\'a}lez et~al.(2025)Ruiz-Gonz{\'a}lez, Rico, and
  Guisasola}]{ruiz2025learning}
\bibinfo{author}{A.~Ruiz-Gonz{\'a}lez}, \bibinfo{author}{A.~Rico},
  \bibinfo{author}{J.~Guisasola},
\newblock \bibinfo{title}{{Learning About Sound in Initial Teacher Training:
  Evaluation and Redesign of a Teaching-Learning Sequence}},
\newblock in: \bibinfo{booktitle}{Connecting Science Education with Cultural
  Heritage: Selected Papers from the ESERA 2023 Conference},
  \bibinfo{organization}{Springer}, \bibinfo{year}{2025}, pp.
  \bibinfo{pages}{157--171}.
\bibitem[{Kleidermacher and Zou(2026)}]{kleidermacher2026science}
\bibinfo{author}{H.~C. Kleidermacher}, \bibinfo{author}{J.~Zou},
\newblock \bibinfo{title}{{Science across languages: assessing LLM multilingual
  translation of scientific papers}},
\newblock in: \bibinfo{booktitle}{Findings of the EACL 2026},
  \bibinfo{year}{2026}, pp. \bibinfo{pages}{3932--3947}.
\bibitem[{Kim et~al.(2024)Kim, Sung, Lee, Lim, and Perez}]{kim2024efficient}
\bibinfo{author}{S.~Kim}, \bibinfo{author}{M.~Sung}, \bibinfo{author}{J.~Lee},
  \bibinfo{author}{H.~Lim}, \bibinfo{author}{J.~G. Perez},
\newblock \bibinfo{title}{{Efficient terminology integration for LLM-based
  translation in specialized domains}},
\newblock in: \bibinfo{booktitle}{Proceedings of the Ninth Conference on
  Machine Translation}, \bibinfo{year}{2024}, pp. \bibinfo{pages}{636--642}.
\bibitem[{Wang et~al.(2023)Wang, Lyu, Ji, Zhang, Yu, Shi, and
  Tu}]{wang2023document}
\bibinfo{author}{L.~Wang}, \bibinfo{author}{C.~Lyu}, \bibinfo{author}{T.~Ji},
  \bibinfo{author}{Z.~Zhang}, \bibinfo{author}{D.~Yu},
  \bibinfo{author}{S.~Shi}, \bibinfo{author}{Z.~Tu},
\newblock \bibinfo{title}{Document-level machine translation with large
  language models},
\newblock in: \bibinfo{booktitle}{Proceedings of the 2023 Conference on
  Empirical Methods in Natural Language Processing}, \bibinfo{year}{2023}, pp.
  \bibinfo{pages}{16646--16661}.
\bibitem[{Zhu et~al.(2024)Zhu, Liu, Dong, Xu, Huang, Kong, Chen, and
  Li}]{zhu2024multilingual}
\bibinfo{author}{W.~Zhu}, \bibinfo{author}{H.~Liu}, \bibinfo{author}{Q.~Dong},
  \bibinfo{author}{J.~Xu}, \bibinfo{author}{S.~Huang},
  \bibinfo{author}{L.~Kong}, \bibinfo{author}{J.~Chen},
  \bibinfo{author}{L.~Li},
\newblock \bibinfo{title}{Multilingual machine translation with large language
  models: Empirical results and analysis},
\newblock in: \bibinfo{booktitle}{Findings of NAACL 2024},
  \bibinfo{year}{2024}, pp. \bibinfo{pages}{2765--2781}.
\bibitem[{Zhang et~al.(2023)Zhang, Haddow, and Birch}]{zhang2023prompting}
\bibinfo{author}{B.~Zhang}, \bibinfo{author}{B.~Haddow},
  \bibinfo{author}{A.~Birch},
\newblock \bibinfo{title}{Prompting large language model for machine
  translation: A case study},
\newblock in: \bibinfo{booktitle}{International conference on machine
  learning}, \bibinfo{organization}{PMLR}, \bibinfo{year}{2023}, pp.
  \bibinfo{pages}{41092--41110}.
\bibitem[{Sabo et~al.(2024)Sabo, Klein, and Bernardinello}]{sabo2024boosting}
\bibinfo{author}{M.~Sabo}, \bibinfo{author}{J.~Klein},
  \bibinfo{author}{G.~Bernardinello},
\newblock \bibinfo{title}{{Boosting machine translation with AI-powered
  terminology features}},
\newblock in: \bibinfo{booktitle}{EAMT}, \bibinfo{year}{2024}, pp.
  \bibinfo{pages}{25--26}.
\bibitem[{Lu et~al.(2024)Lu, Yang, Huang, Zhang, Lam, and Wei}]{lu2024chain}
\bibinfo{author}{H.~Lu}, \bibinfo{author}{H.~Yang}, \bibinfo{author}{H.~Huang},
  \bibinfo{author}{D.~Zhang}, \bibinfo{author}{W.~Lam},
  \bibinfo{author}{F.~Wei},
\newblock \bibinfo{title}{{Chain-of-dictionary prompting elicits translation in
  large language models}},
\newblock in: \bibinfo{booktitle}{EMNLP}, \bibinfo{year}{2024}, pp.
  \bibinfo{pages}{958--976}.
\bibitem[{Kapo{\v{c}}i{\=u}t{\.e}-Dzikien{\.e}
  et~al.(2025)Kapo{\v{c}}i{\=u}t{\.e}-Dzikien{\.e}, Bergmanis, and
  Pinnis}]{kapovciute2025localizing}
\bibinfo{author}{J.~Kapo{\v{c}}i{\=u}t{\.e}-Dzikien{\.e}},
  \bibinfo{author}{T.~Bergmanis}, \bibinfo{author}{M.~Pinnis},
\newblock \bibinfo{title}{{Localizing AI: Evaluating Open-Weight Language
  Models for Languages of Baltic States}},
\newblock in: \bibinfo{booktitle}{NoDaLiDa/Baltic-HLT}, \bibinfo{year}{2025},
  pp. \bibinfo{pages}{287--295}.
\bibitem[{Team et~al.(2024)Team, Mesnard, Hardin, Dadashi, Bhupatiraju, Pathak,
  Sifre, Rivi{\`e}re, Kale, Love et~al.}]{team2024gemma}
\bibinfo{author}{G.~Team}, \bibinfo{author}{T.~Mesnard},
  \bibinfo{author}{C.~Hardin}, \bibinfo{author}{R.~Dadashi},
  \bibinfo{author}{S.~Bhupatiraju}, \bibinfo{author}{S.~Pathak},
  \bibinfo{author}{L.~Sifre}, \bibinfo{author}{M.~Rivi{\`e}re},
  \bibinfo{author}{M.~S. Kale}, \bibinfo{author}{J.~Love}, et~al.,
\newblock \bibinfo{title}{Gemma: Open models based on gemini research and
  technology},
\newblock \bibinfo{journal}{arXiv preprint arXiv:2403.08295}
  (\bibinfo{year}{2024}).
\bibitem[{Grattafiori et~al.(2024)Grattafiori, Dubey, Jauhri, Pandey, Kadian,
  Al-Dahle, Letman, Mathur, Schelten, Vaughan et~al.}]{grattafiori2024llama}
\bibinfo{author}{A.~Grattafiori}, \bibinfo{author}{A.~Dubey},
  \bibinfo{author}{A.~Jauhri}, \bibinfo{author}{A.~Pandey},
  \bibinfo{author}{A.~Kadian}, \bibinfo{author}{A.~Al-Dahle},
  \bibinfo{author}{A.~Letman}, \bibinfo{author}{A.~Mathur},
  \bibinfo{author}{A.~Schelten}, \bibinfo{author}{A.~Vaughan}, et~al.,
\newblock \bibinfo{title}{The llama 3 herd of models},
\newblock \bibinfo{journal}{arXiv preprint arXiv:2407.21783}
  (\bibinfo{year}{2024}).
\bibitem[{Etxaniz et~al.(2024)Etxaniz, Sainz, Miguel, Aldabe, Rigau, Agirre,
  Ormazabal, Artetxe, and Soroa}]{etxaniz2024latxa}
\bibinfo{author}{J.~Etxaniz}, \bibinfo{author}{O.~Sainz},
  \bibinfo{author}{N.~Miguel}, \bibinfo{author}{I.~Aldabe},
  \bibinfo{author}{G.~Rigau}, \bibinfo{author}{E.~Agirre},
  \bibinfo{author}{A.~Ormazabal}, \bibinfo{author}{M.~Artetxe},
  \bibinfo{author}{A.~Soroa},
\newblock \bibinfo{title}{Latxa: An open language model and evaluation suite
  for basque},
\newblock in: \bibinfo{booktitle}{Proceedings of the 62nd Annual Meeting of the
  Association for Computational Linguistics (Volume 1: Long Papers)},
  \bibinfo{year}{2024}, pp. \bibinfo{pages}{14952--14972}.
\bibitem[{Stede et~al.(2019)Stede, Schneider, and
  Hirst}]{stede2019argumentation}
\bibinfo{author}{M.~Stede}, \bibinfo{author}{J.~Schneider},
  \bibinfo{author}{G.~Hirst}, \bibinfo{title}{Argumentation mining},
  \bibinfo{publisher}{Springer}, \bibinfo{year}{2019}.
\bibitem[{Lawrence and Reed(2019)}]{lawrence2019argument}
\bibinfo{author}{J.~Lawrence}, \bibinfo{author}{C.~Reed},
\newblock \bibinfo{title}{Argument mining: A survey},
\newblock \bibinfo{journal}{Computational linguistics} \bibinfo{volume}{45}
  (\bibinfo{year}{2019}) \bibinfo{pages}{765--818}.
\bibitem[{Glockner et~al.(2024)Glockner, Hou, Nakov, and
  Gurevych}]{glockner2024missci}
\bibinfo{author}{M.~Glockner}, \bibinfo{author}{Y.~Hou},
  \bibinfo{author}{P.~Nakov}, \bibinfo{author}{I.~Gurevych},
\newblock \bibinfo{title}{Missci: Reconstructing fallacies in misrepresented
  science},
\newblock in: \bibinfo{booktitle}{Proceedings of the 62nd Annual Meeting of the
  Association for Computational Linguistics (Volume 1: Long Papers)},
  \bibinfo{year}{2024}, pp. \bibinfo{pages}{4372--4405}.
\bibitem[{Augenstein(2021)}]{augenstein2021determining}
\bibinfo{author}{I.~Augenstein},
\newblock \bibinfo{title}{Determining the credibility of science
  communication},
\newblock in: \bibinfo{booktitle}{Proceedings of the Second Workshop on
  Scholarly Document Processing}, \bibinfo{year}{2021}, pp.
  \bibinfo{pages}{1--6}.
\bibitem[{Wright et~al.(2022)Wright, Pei, Jurgens, and
  Augenstein}]{wright2022modeling}
\bibinfo{author}{D.~Wright}, \bibinfo{author}{J.~Pei},
  \bibinfo{author}{D.~Jurgens}, \bibinfo{author}{I.~Augenstein},
\newblock \bibinfo{title}{Modeling information change in science communication
  with semantically matched paraphrases},
\newblock in: \bibinfo{booktitle}{Proceedings of the 2022 Conference on
  Empirical Methods in Natural Language Processing}, \bibinfo{year}{2022}, pp.
  \bibinfo{pages}{1783--1807}.
\bibitem[{Molinet et~al.(2024)Molinet, Marro, Cabrio, and
  Villata}]{molinet2024explanatory}
\bibinfo{author}{B.~Molinet}, \bibinfo{author}{S.~Marro},
  \bibinfo{author}{E.~Cabrio}, \bibinfo{author}{S.~Villata},
\newblock \bibinfo{title}{Explanatory argumentation in natural language for
  correct and incorrect medical diagnoses},
\newblock \bibinfo{journal}{Journal of Biomedical Semantics}
  \bibinfo{volume}{15} (\bibinfo{year}{2024}) \bibinfo{pages}{8}.
\bibitem[{Lemons and Lemons(2013)}]{lemons2013questions}
\bibinfo{author}{P.~P. Lemons}, \bibinfo{author}{J.~D. Lemons},
\newblock \bibinfo{title}{Questions for assessing higher-order cognitive
  skills: It's not just bloom’s},
\newblock \bibinfo{journal}{CBE—Life Sciences Education} \bibinfo{volume}{12}
  (\bibinfo{year}{2013}) \bibinfo{pages}{47--58}.
\bibitem[{Nussbaum(2021)}]{nussbaum2021critical}
\bibinfo{author}{E.~M. Nussbaum},
\newblock \bibinfo{title}{Critical integrative argumentation: Toward complexity
  in students’ thinking},
\newblock \bibinfo{journal}{Educational Psychologist} \bibinfo{volume}{56}
  (\bibinfo{year}{2021}) \bibinfo{pages}{1--17}.
\bibitem[{Gao et~al.(2024)Gao, Merzdorf, Anwar, Hipwell, and
  Srinivasa}]{gao2024automatic}
\bibinfo{author}{R.~Gao}, \bibinfo{author}{H.~E. Merzdorf},
  \bibinfo{author}{S.~Anwar}, \bibinfo{author}{M.~C. Hipwell},
  \bibinfo{author}{A.~R. Srinivasa},
\newblock \bibinfo{title}{Automatic assessment of text-based responses in
  post-secondary education: A systematic review},
\newblock \bibinfo{journal}{Computers and Education: Artificial Intelligence}
  \bibinfo{volume}{6} (\bibinfo{year}{2024}) \bibinfo{pages}{100206}.
\bibitem[{Fu et~al.(2024)Fu, Wei, Hu, Cai, and Liu}]{fu2024qgeval}
\bibinfo{author}{W.~Fu}, \bibinfo{author}{B.~Wei}, \bibinfo{author}{J.~Hu},
  \bibinfo{author}{Z.~Cai}, \bibinfo{author}{J.~Liu},
\newblock \bibinfo{title}{Qgeval: benchmarking multi-dimensional evaluation for
  question generation},
\newblock in: \bibinfo{booktitle}{Proceedings of the 2024 Conference on
  Empirical Methods in Natural Language Processing}, \bibinfo{year}{2024}, pp.
  \bibinfo{pages}{11783--11803}.
\bibitem[{Calvo~Figueras and
  Agerri(2025)}]{calvo-figueras-agerri-2025-benchmarking}
\bibinfo{author}{B.~Calvo~Figueras}, \bibinfo{author}{R.~Agerri},
\newblock \bibinfo{title}{{Benchmarking Critical Questions Generation: A
  Challenging Reasoning Task for Large Language Models}},
\newblock in: \bibinfo{booktitle}{Findings of the EMNLP 2025},
  \bibinfo{year}{2025}, pp. \bibinfo{pages}{5635--5652}.
\bibitem[{Song et~al.(2014)Song, Heilman, Klebanov, and
  Deane}]{song2014applying}
\bibinfo{author}{Y.~Song}, \bibinfo{author}{M.~Heilman}, \bibinfo{author}{B.~B.
  Klebanov}, \bibinfo{author}{P.~Deane},
\newblock \bibinfo{title}{Applying argumentation schemes for essay scoring},
\newblock in: \bibinfo{booktitle}{Proceedings of the first workshop on
  argumentation mining}, \bibinfo{year}{2014}, pp. \bibinfo{pages}{69--78}.
\bibitem[{Nussbaum et~al.(2020)Nussbaum, Tian, Van~Winkle, Perera, Putney,
  Dove, and Carroll}]{nussbaum2020using}
\bibinfo{author}{M.~Nussbaum}, \bibinfo{author}{L.~Tian},
  \bibinfo{author}{M.~Van~Winkle}, \bibinfo{author}{H.~Perera},
  \bibinfo{author}{L.~Putney}, \bibinfo{author}{I.~Dove},
  \bibinfo{author}{K.~Carroll},
\newblock \bibinfo{title}{Using the critical questions model of argumentation
  for science teacher professional learning and student outcomes},
\newblock \bibinfo{journal}{International Society of the Learning Sciences
  (ISLS)}  (\bibinfo{year}{2020}).
\bibitem[{Gu et~al.(2024)Gu, Jiang, Shi, Tan, Zhai, Xu, Li, Shen, Ma, Liu
  et~al.}]{gu2024survey}
\bibinfo{author}{J.~Gu}, \bibinfo{author}{X.~Jiang}, \bibinfo{author}{Z.~Shi},
  \bibinfo{author}{H.~Tan}, \bibinfo{author}{X.~Zhai}, \bibinfo{author}{C.~Xu},
  \bibinfo{author}{W.~Li}, \bibinfo{author}{Y.~Shen}, \bibinfo{author}{S.~Ma},
  \bibinfo{author}{H.~Liu}, et~al.,
\newblock \bibinfo{title}{A survey on llm-as-a-judge},
\newblock \bibinfo{journal}{The Innovation}  (\bibinfo{year}{2024}).
\bibitem[{Li et~al.(2025)Li, Jiang, Huang, Beigi, Zhao, Tan, Bhattacharjee,
  Jiang, Chen, Wu et~al.}]{li2025generation}
\bibinfo{author}{D.~Li}, \bibinfo{author}{B.~Jiang},
  \bibinfo{author}{L.~Huang}, \bibinfo{author}{A.~Beigi},
  \bibinfo{author}{C.~Zhao}, \bibinfo{author}{Z.~Tan},
  \bibinfo{author}{A.~Bhattacharjee}, \bibinfo{author}{Y.~Jiang},
  \bibinfo{author}{C.~Chen}, \bibinfo{author}{T.~Wu}, et~al.,
\newblock \bibinfo{title}{From generation to judgment: Opportunities and
  challenges of llm-as-a-judge},
\newblock in: \bibinfo{booktitle}{Proceedings of the 2025 Conference on
  Empirical Methods in Natural Language Processing}, \bibinfo{year}{2025}, pp.
  \bibinfo{pages}{2757--2791}.
\bibitem[{Alonso et~al.(2024)Alonso, Oronoz, and Agerri}]{alonso2024medexpqa}
\bibinfo{author}{I.~Alonso}, \bibinfo{author}{M.~Oronoz},
  \bibinfo{author}{R.~Agerri},
\newblock \bibinfo{title}{Medexpqa: Multilingual benchmarking of large language
  models for medical question answering},
\newblock \bibinfo{journal}{Artificial intelligence in medicine}
  \bibinfo{volume}{155} (\bibinfo{year}{2024}) \bibinfo{pages}{102938}.
\bibitem[{Garc{\'\i}a-Ferrero et~al.(2024)Garc{\'\i}a-Ferrero, Agerri, Salazar,
  Cabrio, De~la Iglesia, Lavelli, Magnini, Molinet, Ramirez-Romero, Rigau
  et~al.}]{garcia2024medmt5}
\bibinfo{author}{I.~Garc{\'\i}a-Ferrero}, \bibinfo{author}{R.~Agerri},
  \bibinfo{author}{A.~A. Salazar}, \bibinfo{author}{E.~Cabrio},
  \bibinfo{author}{I.~De~la Iglesia}, \bibinfo{author}{A.~Lavelli},
  \bibinfo{author}{B.~Magnini}, \bibinfo{author}{B.~Molinet},
  \bibinfo{author}{J.~Ramirez-Romero}, \bibinfo{author}{G.~Rigau}, et~al.,
\newblock \bibinfo{title}{Medmt5: An open-source multilingual text-to-text llm
  for the medical domain},
\newblock in: \bibinfo{booktitle}{Proceedings of the 2024 Joint International
  Conference on Computational Linguistics, Language Resources and Evaluation
  (LREC-COLING 2024)}, \bibinfo{year}{2024}, pp. \bibinfo{pages}{11165--11177}.
\bibitem[{Sviridova et~al.(2024)Sviridova, Yeginbergen, Estarrona, Cabrio,
  Villata, and Agerri}]{sviridova2024casimedicos}
\bibinfo{author}{E.~Sviridova}, \bibinfo{author}{A.~Yeginbergen},
  \bibinfo{author}{A.~Estarrona}, \bibinfo{author}{E.~Cabrio},
  \bibinfo{author}{S.~Villata}, \bibinfo{author}{R.~Agerri},
\newblock \bibinfo{title}{Casimedicos-arg: A medical question answering dataset
  annotated with explanatory argumentative structures},
\newblock in: \bibinfo{booktitle}{Proceedings of the 2024 conference on
  empirical methods in natural language processing}, \bibinfo{year}{2024}, pp.
  \bibinfo{pages}{18463--18475}.
\bibitem[{Yeginbergen et~al.(2025)Yeginbergen, Oronoz, and
  Agerri}]{yeginbergen2025dynamic}
\bibinfo{author}{A.~Yeginbergen}, \bibinfo{author}{M.~Oronoz},
  \bibinfo{author}{R.~Agerri},
\newblock \bibinfo{title}{Dynamic knowledge integration for evidence-driven
  counter-argument generation with large language models},
\newblock in: \bibinfo{booktitle}{Findings of the Association for Computational
  Linguistics: ACL 2025}, \bibinfo{year}{2025}, pp.
  \bibinfo{pages}{22568--22584}.
\bibitem[{Zubiaga et~al.(2024)Zubiaga, Soroa, and Agerri}]{zubiaga2024llm}
\bibinfo{author}{I.~Zubiaga}, \bibinfo{author}{A.~Soroa},
  \bibinfo{author}{R.~Agerri},
\newblock \bibinfo{title}{A llm-based ranking method for the evaluation of
  automatic counter-narrative generation},
\newblock in: \bibinfo{booktitle}{Findings of the Association for Computational
  Linguistics: EMNLP 2024}, \bibinfo{year}{2024}, pp.
  \bibinfo{pages}{9572--9585}.
\bibitem[{Von~D{\"a}niken et~al.(2024)Von~D{\"a}niken, Deriu, Tuggener, and
  Cieliebak}]{von2024favi}
\bibinfo{author}{P.~Von~D{\"a}niken}, \bibinfo{author}{J.~M. Deriu},
  \bibinfo{author}{D.~Tuggener}, \bibinfo{author}{M.~Cieliebak},
\newblock \bibinfo{title}{Favi-score: A measure for favoritism in automated
  preference ratings for generative ai evaluation},
\newblock in: \bibinfo{booktitle}{Proceedings of the 62nd Annual Meeting of the
  Association for Computational Linguistics (Volume 1: Long Papers)},
  \bibinfo{year}{2024}, pp. \bibinfo{pages}{4437--4454}.
\bibitem[{Deriu et~al.(2023)Deriu, Von~D{\"a}niken, Tuggener, and
  Cieliebak}]{deriu2023correction}
\bibinfo{author}{J.~M. Deriu}, \bibinfo{author}{P.~Von~D{\"a}niken},
  \bibinfo{author}{D.~Tuggener}, \bibinfo{author}{M.~Cieliebak},
\newblock \bibinfo{title}{Correction of errors in preference ratings from
  automated metrics for text generation},
\newblock in: \bibinfo{booktitle}{Findings of the Association for Computational
  Linguistics: ACL 2023}, \bibinfo{year}{2023}, pp.
  \bibinfo{pages}{6456--6474}.
\bibitem[{von D{\"a}niken et~al.(2024)von D{\"a}niken, Deriu, Rodrigo, and
  Cieliebak}]{von2024improving}
\bibinfo{author}{P.~von D{\"a}niken}, \bibinfo{author}{J.~M. Deriu},
  \bibinfo{author}{A.~Rodrigo}, \bibinfo{author}{M.~Cieliebak},
\newblock \bibinfo{title}{Improving quantification with minimal in-domain
  annotations: Beyond classify and count},
\newblock in: \bibinfo{booktitle}{Proceedings of the International AAAI
  Conference on Web and Social Media}, volume~\bibinfo{volume}{18},
  \bibinfo{year}{2024}, pp. \bibinfo{pages}{1585--1598}.
\bibitem[{Zhang et~al.(2024)Zhang, Hu, Zhoubian, Du, Yang, Wang, Yue, Dong, and
  Tang}]{zhang2024sciinstruct}
\bibinfo{author}{D.~Zhang}, \bibinfo{author}{Z.~Hu},
  \bibinfo{author}{S.~Zhoubian}, \bibinfo{author}{Z.~Du},
  \bibinfo{author}{K.~Yang}, \bibinfo{author}{Z.~Wang},
  \bibinfo{author}{Y.~Yue}, \bibinfo{author}{Y.~Dong},
  \bibinfo{author}{J.~Tang},
\newblock \bibinfo{title}{Sciinstruct: a self-reflective instruction annotated
  dataset for training scientific language models},
\newblock \bibinfo{journal}{Advances in Neural Information Processing Systems}
  \bibinfo{volume}{37} (\bibinfo{year}{2024}) \bibinfo{pages}{1443--1473}.
\bibitem[{Xu et~al.(2024)Xu, Jiang, Niu, Deng, Poovendran, Choi, and
  Lin}]{xu2024magpie}
\bibinfo{author}{Z.~Xu}, \bibinfo{author}{F.~Jiang}, \bibinfo{author}{L.~Niu},
  \bibinfo{author}{Y.~Deng}, \bibinfo{author}{R.~Poovendran},
  \bibinfo{author}{Y.~Choi}, \bibinfo{author}{B.~Y. Lin},
\newblock \bibinfo{title}{Magpie: Alignment data synthesis from scratch by
  prompting aligned llms with nothing},
\newblock \bibinfo{journal}{arXiv preprint arXiv:2406.08464}
  (\bibinfo{year}{2024}).
\bibitem[{Bai et~al.(2023)Bai, Bai, Chu, Cui, Dang, Deng, Fan, Ge, Han, Huang
  et~al.}]{bai2023qwen}
\bibinfo{author}{J.~Bai}, \bibinfo{author}{S.~Bai}, \bibinfo{author}{Y.~Chu},
  \bibinfo{author}{Z.~Cui}, \bibinfo{author}{K.~Dang},
  \bibinfo{author}{X.~Deng}, \bibinfo{author}{Y.~Fan}, \bibinfo{author}{W.~Ge},
  \bibinfo{author}{Y.~Han}, \bibinfo{author}{F.~Huang}, et~al.,
\newblock \bibinfo{title}{Qwen technical report},
\newblock \bibinfo{journal}{arXiv preprint arXiv:2309.16609}
  (\bibinfo{year}{2023}).

\end{thebibliography}


\end{document}